\documentclass[lettersize,journal]{IEEEtran}
\pdfoutput=1
\usepackage{amsmath,amsfonts}
\usepackage{algorithmic}
\usepackage{algorithm}
\usepackage{array}
\usepackage{textcomp}
\usepackage{stfloats}
\usepackage{url}
\usepackage{verbatim}
\usepackage{graphicx}
\usepackage{cite}
\usepackage{booktabs}
\usepackage{multirow}
\usepackage{makecell}
\usepackage{stackengine}
\usepackage{threeparttable}
\usepackage{xcolor} 
\usepackage{tikz}
\usepackage{soul}
\usepackage[normalem]{ulem}
\usepackage{diagbox}
\usepackage{subfigure}

\usepackage[colorlinks,
bookmarksopen,
bookmarksnumbered,
citecolor=blue,
linkcolor=red,
urlcolor=red]{hyperref}
\usepackage{bookmark}

\hypersetup{pdfborder={0 0 0}}

\hypersetup{hidelinks,
	colorlinks=true,
	allcolors=black,
	pdfstartview=Fit,
	breaklinks=true}
	
\definecolor{lime}{HTML}{A6CE39}

\DeclareRobustCommand{\orcidicon}{
\begin{tikzpicture}
\draw[lime, fill=lime] (0,0)
circle[radius=0.16]
node[white]{{\fontfamily{qag}\selectfont \tiny \.{I}D}};
\end{tikzpicture}
\hspace{-2mm}
}
\foreach \x in {A, ..., Z}{%
\expandafter\xdef\csname orcid\x\endcsname{\noexpand\href{https://orcid.org/\csname orcidauthor\x\endcsname}{\noexpand\orcidicon}}
}

\begin{document}

\title{SELC: Self-Supervised Efficient Local Correspondence Learning for Low Quality Images}

\author{Yuqing Wang\hspace{-1.5mm} , Yan Wang\hspace{-1.5mm} , Hailiang Tang\hspace{-1.5mm} , Xiaoji Niu\hspace{-1.5mm} , ~\IEEEmembership{Member,~IEEE}
\thanks{This paper was produced by the IEEE Publication Technology Group. They are in Piscataway, NJ.}
\thanks{Manuscript received April 19, 2021; revised August 16, 2021.}}

\markboth{Journal of \LaTeX\ Class Files,~Vol.~14, No.~8, August~2021}%
{Shell \MakeLowercase{\textit{et al.}}: A Sample Article Using IEEEtran.cls for IEEE Journals}

\IEEEpubid{}

\maketitle

\begin{abstract}
Accurate and stable feature matching is critical for computer vision tasks, particularly in applications such as Simultaneous Localization and Mapping (SLAM). While recent learning-based feature matching methods have demonstrated promising performance in challenging spatiotemporal scenarios, they still face inherent trade-offs between accuracy and computational efficiency in specific settings. In this paper, we propose a lightweight feature matching network designed to establish sparse, stable, and consistent correspondence between multiple frames. The proposed method eliminates the dependency on manual annotations during training and mitigates feature drift through a hybrid self-supervised paradigm. Extensive experiments validate three key advantages: (1)  Our method operates without dependency on external prior knowledge and seamlessly incorporates its hybrid training mechanism into original datasets. (2) Benchmarked against state-of-the-art deep learning-based methods, our approach maintains equivalent computational efficiency at low-resolution scales while achieving a 2-10× improvement in computational efficiency for high-resolution inputs. (3) Comparative evaluations demonstrate that the proposed hybrid self-supervised scheme effectively mitigates feature drift in long-term tracking while maintaining consistent representation across image sequences.
\end{abstract}

\begin{IEEEkeywords}
Visual localization, Feature matching, Lightning network, Single \& Multiple Consistency.
\end{IEEEkeywords}

\section{Introduction}\label{I}

\IEEEPARstart{I}{mage} feature tracking constitutes a fundamental challenge in 3D computer vision, as establishing reliable correspondences serves as a prerequisite for critical downstream applications including Structure from Motion (SfM) \cite{7780814}, visual localization \cite{7898369,6906584}, and Simultaneous Localization and Mapping (SLAM) \cite{8421746,chen2022aspanformer}. While recent advances \cite{Edstedt_2023_CVPR, sun2021loftr,10054148} have demonstrated progressive improvements in computational efficiency for low-resolution imagery and short-term matching accuracy, these achievements frequently occur at the expense of network generalizability, long-term tracking precision, and processing efficiency for high-resolution inputs.

Existing partial matching paradigms exhibit strong dependency on manually annotated supervision signals during training \cite{Iscen_2019_CVPR, csurka2017domainadaptationvisualapplications}, where pose priors and manually labeled depth maps are fused into high-dimensional feature descriptor learning. Although such annotations provide rich supervisory signals, this supervision strategy fails to comprehensively encapsulate real-world degraded scenarios while introducing increased computational overhead. Furthermore, inherent inaccuracies in manual annotations within motion-blurred regions propagate through the learning process, inducing subtle discrepancies in identical feature point descriptors. To address annotation-related limitations, researchers have proposed integrating feature tracking pipelines with selection mechanisms \cite{9726928,10556560} that enhance tracking quality by avoiding keypoint sampling in high-ambiguity regions. However, keypoint selection quality alone cannot resolve persistent challenges arising from occlusion, motion blur, and illumination variations during dynamic movements. These factors progressively amplify error accumulation in network inference, potentially leading to catastrophic accuracy degradation if not systematically addressed \cite{10.5555/3045390.3045643}.

Notably, a fundamental dichotomy exists between feature extraction paradigms emphasizing global information utilization and long-term tracking requirements demanding evolutionary analysis of local patterns. This inherent contradiction emphasizes the necessity for effective exploitation of local feature dynamics to achieve persistent, efficient, and stable tracking - a critical requirement for SLAM systems and analogous downstream applications.

In this paper, we propose a complete and robust lightweight local descriptor hybrid self-supervision method to systematically address these two problems. The inherent sparsity of local feature similarity is exploited to implement a lightweight fully convolutional neural network (CNN) architecture. A dual matching mechanism is proposed to ensure the consistency of feature point description by leveraging the consistency of local features between a single frame and consecutive frames. To avoid the unpredictability introduced by manual annotation information, we develop an auxiliary guidance strategy to provide reliable supervision signals. Our framework innovatively combines traditional tracking methods with a CNN architecture, where handcrafted descriptors (e.g., ORB) and photometric constraints (e.g., KLT) are systematically incorporated as supervision signals.

In summary, we present a prior-free lightweight CNN-based feature tracking network that seamlessly integrates with existing SLAM pipelines. Our work makes three pivotal contributions:

\begin{itemize}
\item{We develop a computationally efficient CNN framework that ensures stable feature correspondence through low-dimensional descriptors and local patch information. This framework is specifically designed for deployment on resource-constrained platforms without requiring hardware-specific optimizations, making it suitable for high-throughput applications.}
\item{We propose an auxiliary network training strategy that leverages traditional schemes, eliminating the need for manual annotations while enabling seamless integration with any SLAM system.}
\item{We design a hybrid self-supervision paradigm that effectively mitigates tracking drift while maintaining sub-pixel accuracy under challenging imaging conditions by enforcing multi-frame consistency.}

\end{itemize}

\section{Related Work}\label{II}

Modern image matching techniques include both classical and deep learning-based keypoint detection and matching methods. Classical methods, such as the Harris corner detector \cite{123456} and ORB \cite{6126544}, utilize multi-scale pyramids and orientation assignment steps for keypoint detection and description. These methods store and match high-dimensional descriptors, which require substantial computational resources. Additionally, the limited number of descriptors may lack robustness to noise and invariance to scale and rotation.

On the other hand, deep learning-based methods can be categorized into global and local matching approaches. Global matching methods, such as SEKD \cite{song_sekd}, propose a self-supervised framework for learning high-level local features from unlabeled natural images, while DALF \cite{10204558} introduces a novel deformation-aware network for joint keypoint detection and description, designed to address challenging deformable surface matching tasks. However, these methods still require high-dimensional descriptors to be extracted from global context, leading to high computational costs, especially when applied to high-resolution images in matching tasks. In contrast, block-based local matching methods estimate descriptor similarity from image patches and often employ cross-entropy \cite{7298948} or triplet loss \cite{inproceedings} for training. By focusing on information within a specific pixel block, these methods significantly reduce computational costs and have been widely adopted in subsequent works \cite{10.5555/3295222.3295236,8575521,8954064}. However, block-based methods focus solely on descriptor extraction, and their receptive fields are limited to the image patch.

Recently, mid-level approaches, referred to as learned matchers \cite{8954064,10550828,10377620}, as well as end-to-end semi-dense \cite{Sarlin_2020_CVPR,chen2022aspanformer} and dense \cite{sun2021loftr,Edstedt_2023_CVPR, vilain2024semidensedetectorfreemethodsgood} methods, have demonstrated significant improvements in robustness and accuracy when matching wide-baseline image pairs. These advances are particularly notable with the latest developments in Transformer architectures \cite{10054148}. However, recent methods predominantly emphasize accuracy and robustness in image matching, resulting in inflated computational demands, which are less than ideal, even for systems equipped with moderate GPU resources. Significant modifications are required for these methods to effectively operate in large-scale downstream tasks, such as visual localization, simultaneous localization and mapping (SLAM), and structure-from-motion (SfM).

To maintain low computational requirements, local feature extraction remains the standard approach for key tasks such as image matching and retrieval. Our method is inspired by block-based and traditional descriptor-free optical flow tracking techniques, leveraging conventional optical flow to warm up the network and enable basic feature tracking capabilities.

In Loftr, feature correlation arises from global image representations, which may be redundant in SLAM systems with limited motion dynamics. To address this and considering the minimal capacity of CNN architectures, we restrict the correlation within local patches, ensuring that any patch containing a specific keypoint exhibits consistent properties across the entire network, manifesting as intra-frame consistency. Additionally, as emphasized in \cite{10.5555/3295222.3295349}, the quality of matching needs to be evaluated through proxy tasks. Therefore, for SLAM, it is essential to establish an evaluation metric that links pose estimation quality with feature matching. We enforce inter-frame consistency for any given keypoint, ensuring that patches containing the same keypoint across multiple images exhibit similar properties. This design not only maintains stable intra-frame consistency but also enables strong inter-frame consistency for SLAM applications.

Compared to existing methods, our approach focuses on efficient and stable continuous image matching, utilizing a single backbone network for subsequent learning, making it feasible to deploy SLAM solutions on low-cost motherboards and embedded systems.

\section{Method}\label{III}

\subsection{Preprocessing}\label{III-A}
Given a pair of images $\{\mathbf{I}_A, \mathbf{I}_B\}$, our objective is to obtain a set of 2D correspondences, represented as $\{(\mathbf{p}_A^n, \mathbf{p}_B^n)\}_{n=1}^{N}$. First, a feature detector (e.g., SuperPoint detector) is applied to image $\mathbf{I}_A$ to extract a set of $N$ keypoints, $\mathbf{p}_A^n$. Then, the proposed network is used to identify the most similar feature points $\mathbf{p}_B^n$ in image $\mathbf{I}_B$ by maximizing similarity. In this context, the feature matching problem is simplified to a sparse-to-dense matching problem, i.e., finding the corresponding $\mathbf{p}_B^n$ in image $\mathbf{I}_B$ for each detected $\mathbf{p}_A^n$. We propose to convert this correspondence learning problem into supervised and unsupervised tasks by restricting the set of acceptable positions to the pixel coordinates of $\mathbf{I}_B$.

To eliminate the reliance on manual annotation information during training, we combine traditional tracking methods with deep learning. The bidirectional Lucas-Kanade (LK) optical flow is applied to track feature points $\mathbf{p}_A^n$, filtering out those that cannot be successfully tracked. The filtered points are denoted as $\tilde{\mathbf{p}}_A^n$, and their corresponding points in image $\mathbf{I}_B$ are denoted as $\tilde{\mathbf{p}}_B^n$. These filtered points serve as ground truth coordinates, laying the foundation for constructing subsequent heatmaps and probability map constraints.

High-resolution images contain a wealth of details; therefore, performing full-image convolutions often blurs these details, especially during downsampling. To address this issue, we utilize local image regions to gather the necessary information. For each $\tilde{\mathbf{p}}_A^n$ and $\tilde{\mathbf{p}}_B^n$, we extract $n$ image patches of fixed size $W_A \times H_A \times \text{dim}$ and $W_B \times H_B \times \text{dim}$ from images $\mathbf{I}_A$ and $\mathbf{I}_B$, respectively. These image patches are denoted as $\mathbf{H}_A^n$ and $\mathbf{H}_B^n$. Notably, if the input images are grayscale, the feature map's dimension size is $1$, while for color images, it is $3$. Each patch is processed individually to address the computational complexity caused by high-resolution images.

For the long-term tracking drift issues, we adopt hybrid self-supervised single-frame and multi-frame consistency. In the hybrid self-supervised phase, we introduce random offsets for each patch and repeat the extraction process $M_1$ times to expand the receptive field and enhance intra-frame consistency. Finally, we extract $M_2$ image patches from consecutive tracking frames to strengthen inter-frame consistency constraints.

This approach ensures accurate local feature matching while preserving rich details during feature extraction, which is crucial for robust matching performance in high-resolution images.

\subsection{Network architecture}\label{III-B}

\begin{figure*}
	\centering
	{\includegraphics[width=1\linewidth]{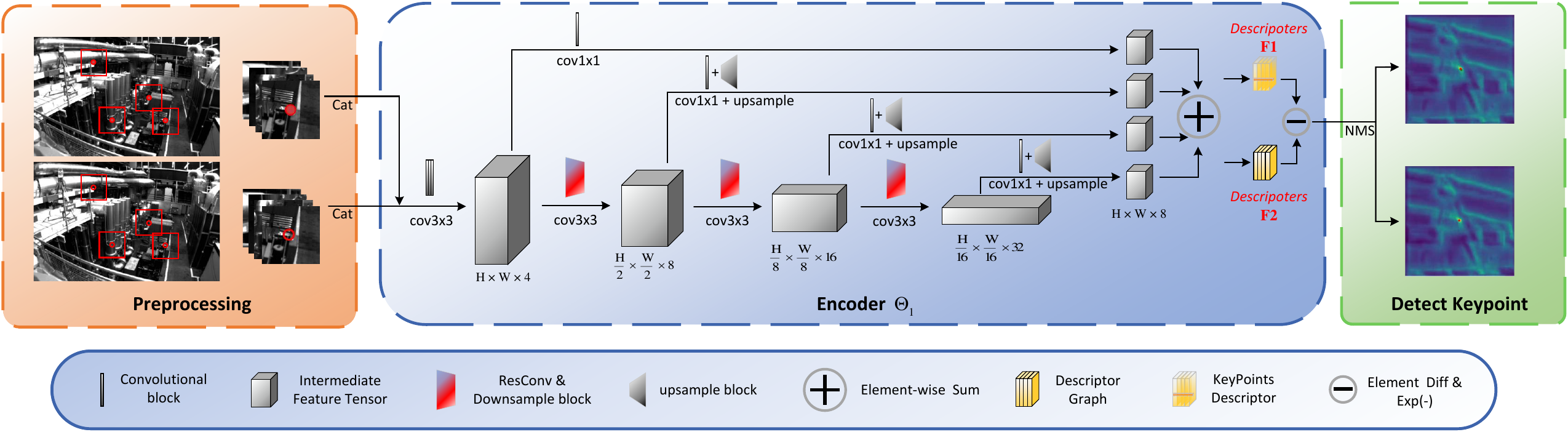}}
	
	\caption{\textbf{The pipeline of the proposed method}. In the preprocessing stage, fixed-size image patches surrounding the keypoints are extracted. The method achieves exceptional speed through shallow convolutional operations, followed by the generation of a compact 32-dimensional dense descriptor map $\mathbf{D}$ in the subsequent encoding phase. Sub-pixel feature locations are obtained through similarity computation and differentiable feature extraction}
	\label{Fig:2}
\end{figure*}

To achieve a lightweight network and faster inference speed, our network consists of one fully convolutional networks, $\Theta_1$, as shown in Fig. \ref{Fig:2}. $\Theta_1$ comprises four fully convolutional layers, normalization layers, activation functions, and pooling layers.

The image patches extracted during preprocessing are fed into $\Theta_1$, generating an intermediate feature descriptor map of size $W_A \times H_A \times 4$ after the first convolutional layers.  Subsequently, the feature maps $\mathbf{F}_A$ and $\mathbf{F}_B$ are passed into $\Theta_2$, which processes them through three convolutional layers and a pooling layer to generate intermediate multi-scale feature descriptor maps of varying dimensions. Finally, all feature descriptor maps are upsampled and concatenated to form the dense feature descriptor maps $\mathbf{D}_A$ and $\mathbf{D}_B \in \mathbb{R}^{W_A \times H_A \times \operatorname{dim}}$ for each patch. Such a design can focus on the local information of the image block while ensuring a low feature dimension, making it easier to obtain the subsequent low-dimensional feature point descriptors.

Using bilinear interpolation, the dense descriptors $\mathbf{d}_{\tilde{\mathbf{p}}_A}$ and $\mathbf{d}_{\tilde{\mathbf{p}}_B}\in \mathbb{R}^{\operatorname{dim}}$ are obtained for the corresponding feature points $\tilde{\mathbf{p}}_A$ and $\tilde{\mathbf{p}}_B$.

\subsection{Supervise Loss}\label{III-C}

Compared with existing sparse-to-sparse matching paradigms (e.g., R2D2 \cite{10.5555/3454287.3455400}) and sparse-to-dense approaches (e.g., S2DNet \cite{germain2020s2dnetlearningaccuratecorrespondences}), our method integrates multiple training paradigms to accelerate convergence. Specifically, ground-truth optical flow tracking is employed to establish sparse correspondences, while deep feature maps  $\mathbf{D}$ combined with keypoint descriptors $\mathbf{d}_{\tilde{\mathbf{p}}}$  are utilized to construct three complementary loss functions: peak loss for salient feature enhancement, descriptor heatmap loss for spatial distribution regularization, and probabilistic dense descriptor loss for distribution alignment.

\subsubsection{Keypoint loss}\label{III-C(1)}
Once the dense descriptors $\mathbf{d}_{\tilde{\mathbf{p}}_A}$ and $\mathbf{d}_{\tilde{\mathbf{p}}_B}$ for the keypoints are obtained, we can use Einstein summation notation to construct associations between them and the dense descriptor maps $\mathbf{D}_B$ and $\mathbf{D}_A$, resulting in similarity maps $\mathbf{C}^{\mathbf{d}_{\tilde{\mathbf{p}}_A}}_{\mathbf{D}_B}$ and $\mathbf{C}^{\mathbf{d}_{\tilde{\mathbf{p}}_B}}_{\mathbf{D}_A}$ of size $W_A \times H_A \times dim$ as
\begin{equation}
\mathbf{C}^{\mathbf{d}_{\tilde{\mathbf{p}}_B}}_{\mathbf{D}_A}=\mathbf{D}_B \mathbf{d}_{\tilde{\mathbf{p}}_A}
\label{1}
\end{equation}

By applying non-maximum suppression (NMS) on the similarity maps, we obtain the mapped keypoints $\tilde{\mathbf{p}}_A$ in $\mathbf{H}_B$, denoted as ${}^{\mathbf{H}_B}\tilde{\mathbf{p}}_A$, where the superscript indicates the original point, and the subscript denotes the mapped image.

As depicted in Fig. \ref{Fig:3}, the reprojection error is defined as the distance between the points of maximum similarity and the points tracked by optical flow

\begin{equation}
\operatorname{dist}(\tilde{\mathbf{p}}_A^{(i)}, \tilde{\mathbf{p}}_B^{(i)})=\left\| {}^{\mathbf{H}_B}\tilde{\mathbf{p}}^{(i)}_A-\tilde{\mathbf{p}}^{(i)}_B\right\|
\label{2}
\end{equation}
where $\|\cdot\|$ is $l2$-norm of the vector and $\operatorname{dist}^{\tilde{\mathbf{p}}_A}_{\tilde{\mathbf{p}}_B}$ is the distance from the point $\tilde{\mathbf{p}}_A$ on $\mathbf{H}_A$ to the point $\tilde{\mathbf{p}}_B$ on $\mathbf{H}_B$. Similar to bidirectional optical flow, we define a symmetric form of reprojection residual as 

\begin{equation}
	{\mathcal{L}_{rp}} = \frac{1}{N} \sum_{i=0}^{N-1} \left( \operatorname{dist}(\tilde{\mathbf{p}}_A^{(i)}, \tilde{\mathbf{p}}_B^{(i)}) + \operatorname{dist}(\tilde{\mathbf{p}}_B^{(i)}, \tilde{\mathbf{p}}_A^{(i)}) \right)
	\label{3}
\end{equation}
where $N$ is the number of patches extracted from image $\mathbf{I}_A$.

\subsubsection{Local peaky loss}\label{III-C(2)}

In the similarity map $C$of each patch, there should be exactly one pixel with the highest similarity score to the keypoint. However, around this pixel, there may be many other pixels with similarity scores that are lying between the highest score and the average score. 

To ensure that the keypoint forms a peak within its neighborhood, we adopt the peak-maximization strategy as discussed in \cite{10556560}. Specifically, consider a N × N -sized patch near the keypoint $\tilde{\mathbf{p}}_A$ on the similarity map, the distance from each pixel at position $[i^{'},j^{'}]$ to the keypoint $\tilde{\mathbf{p}}_A$ is

\begin{equation}
	d(\tilde{\mathbf{p}}_A, i^{'}, j^{'})=\left\{\left\|\tilde{\mathbf{p}}_A-[i^{'},j^{'}]\right\| \mid 0 \leq i^{'}, j^{'}<N\right\}
	\label{4}
\end{equation}

Distance can be utilized as a weighting factor for similarity scores to enhance the discriminability of feature points within local regions. Initially, the scores are normalized to the range of 0 to 1 using the softmax function
\begin{equation}
	s^{\prime}(i^{'}, j^{'})=\operatorname{softmax}\left(\frac{s(i^{'}, j^{'})-s_{\max }}{t_{\operatorname{det}}}\right)
	\label{5}
\end{equation}
where $t_{\operatorname{det}}$ is the temperature which controls the “sharpness” of the normalization, $s(i^{'}, j^{'})$represents the similarity score of pixel $[i^{'}, j^{'}]$.

Subsequently, the local peaky loss of $\tilde{\mathbf{p}}_A$ can be defined as
\begin{equation}
	\mathcal{L}_{lpkA}=\frac{1}{N^2} \sum_{0 \leq i^{'}, j^{'}<N} d(\tilde{\mathbf{p}}_A, i^{'}, j^{'}) s^{\prime}(i^{'}, j^{'})
	\label{6}
\end{equation}

The local peaky loss can be defined as
\begin{equation}
\mathcal{L}_{lpk}=\mathcal{L}_{lpkA}+\mathcal{L}_{lpkB}
\label{7}
\end{equation}

\subsubsection{Heat map loss}\label{III-C(3)}

The local maxima generated by the peak loss are insufficient, as they do not impose constraints on areas outside the neighborhood of candidate points. We extend the concept of peak loss to the entire patch. 

Since there is exactly one feature candidate point with the highest similarity within each patch, we can construct a similarity heatmap $\mathbf{G}^{\tilde{\mathbf{p}}_B}_t$ with the same size of $\mathbf{C}^{\mathbf{d}_{\tilde{\mathbf{p}}_B}}_{\mathbf{D}_A}$ based on a Gaussian distribution using the pre-tracked ground truth $\tilde{\mathbf{p}}_B$
\begin{equation}
	\mathbf{G}(x, y)=\frac{1}{2 \pi \sigma^2} \exp \left(-\frac{\left(x-x_B\right)^2+\left(y-y_B\right)^2}{2 \sigma^2}\right)
	\label{8}
\end{equation}
where $(x_0, y_0$) is the coordinate of $\tilde{\mathbf{p}}_B$.

Then the heat map loss of $\mathbf{C}^{\mathbf{d}_{\tilde{\mathbf{p}}_B}}_{\mathbf{D}_A}$ and $\mathbf{G}^{\tilde{\mathbf{p}}_B}_t$ can be defined as 
\begin{equation}
	\mathcal{L}_{hmB}=\left\| \mathbf{C}^{\mathbf{d}_{\tilde{\mathbf{p}}_B}}_{\mathbf{D}_A}-\mathbf{G}^{\tilde{\mathbf{p}}_B}_t\right\|_{mse}
	\label{9}
\end{equation}
where the  $\|\cdot\|_{mse}$ is MSE-lose of vector.The heat map loss can be defined as
\begin{equation}
	\mathcal{L}_{hm}=\mathcal{L}_{hmB}+\mathcal{L}_{hmA}
	\label{10}
\end{equation}

\subsubsection{Dense descriptor loss}\label{III-C(4)}
As described in [24], the NRE function is used to learn descriptors, and it has been widely applied in subsequent works. The core idea of this method is incorporated into our approach, where the similarity map after a softmax activation represents the similarity of each point as a probability distribution. The tracked feature points are expected to correspond to the highest probability values. The matching probability map is defined as 
${\mathbf{P}^{\mathbf{d}_{\tilde{\mathbf{p}}_A}}_{\mathbf{D}_B}}\in \mathbb{R}^{H \times W}$
\begin{equation}
\mathbf{P}^{\mathbf{d}_{\tilde{\mathbf{p}}_A}}_{\mathbf{D}_B}=\operatorname{softmax}\left(\frac{\mathbf{C}^{\mathbf{d}_{\tilde{\mathbf{p}}_A}}_{\mathbf{D}_B} -1}{t}\right)
\label{11}
\end{equation}
where the normalization function softmax converts similarity to probability and satisfies that all elements sum to one, i.e., $\sum_{H \times W} {\mathbf{P}^{\mathbf{d}_{\tilde{\mathbf{p}}_A}}_{\mathbf{D}_B}}\in \mathbb{R}^{H \times W} = 1$. For the tracked point $\tilde{\mathbf{p}}_A$, its tracking result in $\mathbf{H}_B$ is defined as ${}^{\mathbf{H}_B}\tilde{\mathbf{p}}_A$, and its probability value ${}^{\mathbf{H}_B}\mathbf{p}_A$ can be obtained by bilinear interpolation in the similarity map ${\mathbf{P}^{\mathbf{d}_{\tilde{\mathbf{p}}_A}}_{\mathbf{D}_B}}$
\begin{equation}
{}^{\mathbf{H}_B}\mathbf{p}_A= bisampling \left({\mathbf{P}^{\mathbf{d}_{\tilde{\mathbf{p}}_A}}_{\mathbf{D}_B}}, {}^{\mathbf{H}_B}\tilde{\mathbf{p}}_A\right)
\label{12}
\end{equation}

Maximizing the matching probability at the projection location ${}^{\mathbf{H}_B}\mathbf{p}_A$
subject to the constraint that the sum of all elements equals 1 implies minimizing the negative logarithm of the probability at that point. The descriptor loss function is then obtained as
\begin{equation}
\mathcal{L}_{desc}=\frac{1}{2} \cdot( -\ln ({{}^{\mathbf{H}_B}\mathbf{p}_A})- \ln ({{}^{\mathbf{H}_A}\mathbf{p}_B}))
\label{13}
\end{equation}
where${{}^{\mathbf{H}_A}\mathbf{p}_B}$ means the keypoint in $\mathbf{H}_A$, and $-\ln (\cdot)$ function converts the maximization problem into a minimization problem.

The final supervise loss $\mathcal{L}_{sup}$ is then a linear combination of all losses  
\begin{equation}
\mathcal{L}_{sup}=\alpha \mathcal{L}_{rp}+\beta \mathcal{L}_{lpk}+\gamma \mathcal{L}_{hm}+\delta \mathcal{L}_{desc}
\label{14}
\end{equation}

\subsection{Single \& Multiple Consistency Loss}\label{III-D}
Unlike existing open-source methods that rely on depth maps and pose priors for training, our method is designed to track arbitrary feature points in any scenario without prior information. Therefore, it is crucial to fully exploit both intra-frame correlations within a single image and inter-frame correlations across consecutive images. These correlations are systematically integrated into our framework to establish both single-frame and multi-frame constraints.

\begin{figure}
	\centering
	{\includegraphics[width=0.9\linewidth]{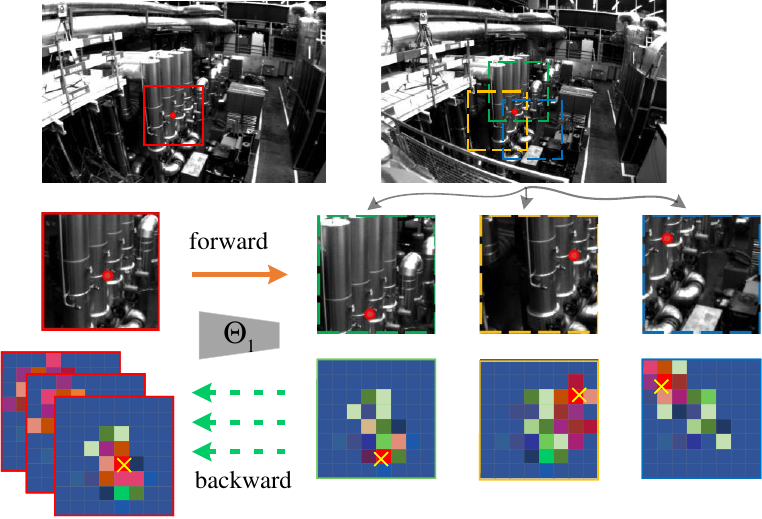}}
	\caption{\textbf{Single Consistency Loss}. The red dots represent the matched features. For each tracked point in the first frame, corresponding image patches at different locations of the ground truth in the second frame are identified. A cost function is constructed using the coordinates $\tilde{\mathbf{p}}^i$ of multiple randomly shifted image patches relative to each other.}
	\label{Fig:3}
\end{figure}

\subsubsection{Single consistency loss}\label{III-D(1)}
The network trained with the supervision signals described in Sec.\ref{III-C} has acquired the capability to track feature points over short time intervals as shown in Fig. \ref{Fig:3}. However, due to the limited size of local image patches, the network is unable to perceive information beyond patch boundaries. This limitation hinders the establishment of consistent constraints when dealing with repetitive textures or rapidly changing dynamic backgrounds. To effectively mitigate the limitations introduced by patch-based feature tracking, we introduce an unsupervised consistency constraint within the same feature point, referred to as the inter-frame constraint.

Firstly, $m$ and $n$ patches are randomly extracted around the feature point $\tilde{\mathbf{p}}_A$ in $\mathbf{I}_A$ and the feature point $\tilde{\mathbf{p}}_B$ in $\mathbf{I}_B$, respectively, denoted as $\mathbf{H}^{m}_A$ and $\mathbf{H}^{n}_B$. Subsequently, each pair $\{\mathbf{H}^{m}_A, \mathbf{H}^{n}_B\}$ is processed to infer the tracking results for each pair, represented as $\{{}^{\mathbf{H}^{j}_B}\tilde{\mathbf{p}}^{i}_A,{}^{\mathbf{H}^{i}_A}\tilde{\mathbf{p}}^{j}_B \}$, where $i \in m, j \in n $.

The single re-projection consistency loss among these results can be defined as 

\begin{align}
	\mathcal{L}_{srp} = \frac{1}{m+n} \sum_{\substack{i,j=0}}^{\substack{m-1 \\ n-1}} \big(
	& \operatorname{dist_{th}}({}^{\mathbf{H}^{j}_B}\tilde{\mathbf{p}}_A^{(i)}, {}^{\mathbf{H}^{j}_B}\tilde{\mathbf{p}}_A^{(i+1)}) + \notag \\
	& \operatorname{dist_{th}}({}^{\mathbf{H}^{i}_A}\tilde{\mathbf{p}}^{(j)}_B, {}^{\mathbf{H}^{i}_A}\tilde{\mathbf{p}}^{(j+1)}_B) \big)
	\label{15}
\end{align}

where the $\operatorname{dist}(\cdot)$ is defined as \eqref{2}, note that all feature points generated in this step should be scaled to their pixel positions in the original image, rather than constructing the loss within the local patches. This ensures that the loss computation is performed at the original image resolution, maintaining spatial consistency and accuracy in the optimization process.

\begin{figure*}
	\centering
	{\includegraphics[width=0.9\linewidth]{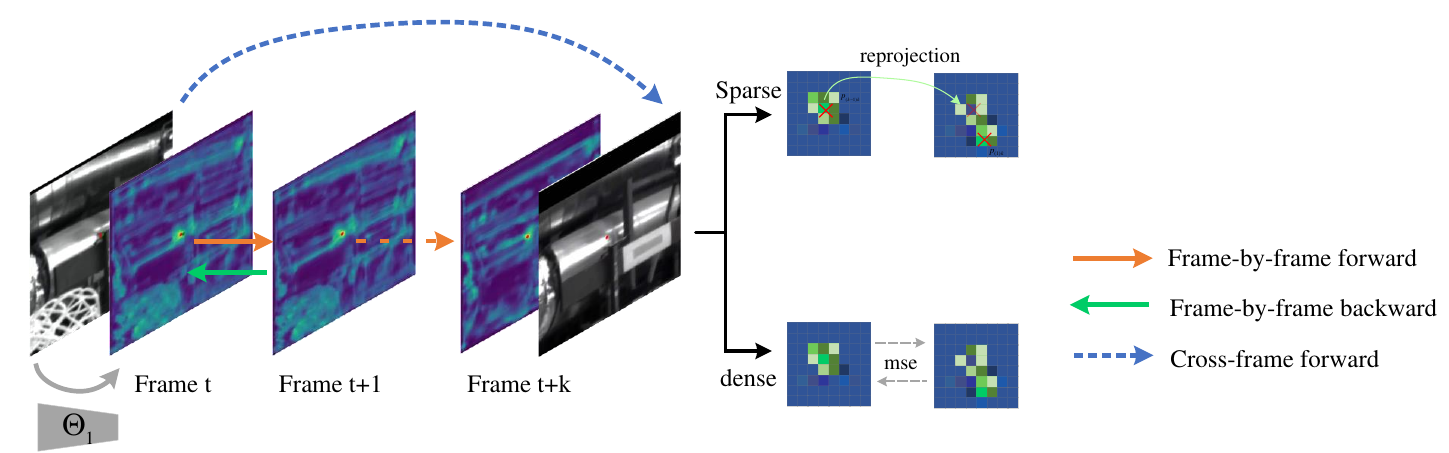}}
	\caption{\textbf{Multiple Consistency Loss}. The red points represent positions obtained from bidirectional optical flow, while the green points denote target tracking points. Utilizing the feature maps over multiple epochs, the cost function is computed between the frame-by-frame extracted feature point coordinates $\tilde{\mathbf{p}}$ and similarity map $\mathbf{C}$, as well as the cross-frame extracted feature point coordinates $\tilde{\mathbf{p}}'$ and similarity map $\mathbf{C}'$.}
	\label{Fig:4}
\end{figure*}

\subsubsection{Multiple consistency loss}\label{III-D(2)}
For downstream tasks such as visual odometry and 3D reconstruction, ensuring the consistency of feature points over long-term tracking is crucial. Specifically, in long sequences, tracking may gradually deviate from the true trajectory due to accumulated errors (e.g., tracking drift), while local patches struggle to capture motion patterns across multiple frames. To address this issue, the proposed method incorporates a multi-frame constraint, which combines dense self-supervision with sparse self-supervision to enforce consistency across multiple frames, as illustrated in Fig. \ref{Fig:4}.

Firstly, feature points $\mathbf{p}_m$ that can be continuously tracked for more than three frames in a long video sequence $\mathbf{M}$ are extracted. For each feature point $\mathbf{p}_i \in \mathbf{p}_m$, the corresponding patches in $\mathbf{M}$ are denoted as $\mathbf{H}^{\mathbf{p}_i}_{\mathbf{M}_j}$ $( \mathbf{M}_j \in \mathbf{M} )$. The point propagated sequentially from the first frame to frame $\mathbf{M}_j$ is defined as ${}^{\mathbf{H}^{p_i}_{\mathbf{M}_j}}\tilde{\mathbf{p}}_{1->\mathbf{M}}$, while the point directly estimated from the first frame to frame $\mathbf{M}_j$ is defined as ${}^{\mathbf{H}^{\mathbf{p}_i}_{\mathbf{M}_j}}\tilde{\mathbf{p}}_{1=>\mathbf{M}}$. To enhance multi-frame consistency, its sparse self-supervised paradigm can be formulated as
\begin{align}
	\mathcal{L}_{mrp} = \frac{1}{m} \sum_{i=0}^{m-1} \big( 
	& \operatorname{dist_{th}}({}^{\mathbf{H}^{p_i}_{\mathbf{M}_j}}\tilde{\mathbf{p}}_{1->\mathbf{M}}, {}^{\mathbf{H}^{\mathbf{p}_i}_{\mathbf{M}_j}}\tilde{\mathbf{p}}_{1=>\mathbf{M}}))
	\label{16}
\end{align}

Secondly, to ensure the consistency of dense self-supervision, the consistency of the similarity map is also taken into account. The method for computing the similarity map is introduced in Sec.\ref{III-C(1)}. We define the dense descriptor map of frame $M_i$ as $\mathbf{D}_{M_i}$, and the feature point descriptor of frame $M_i - 1$ as $ \mathbf{d}_{\tilde{\mathbf{p}}_{M_i-1}}$. Therefore, the similarity map is given by $\mathbf{C}^{\mathbf{d}_{\tilde{\mathbf{p}}_{M_i-1}}}_{\mathbf{D}_{M_i}}$. Similarly, the feature point descriptor of the first frame is defined as $\mathbf{d}_{\tilde{\mathbf{p}}_1}$, and its corresponding similarity map is $\mathbf{C}^{\mathbf{d}_{\tilde{\mathbf{p}}_1}}_{\mathbf{D}_{M_i}}$. Thus, the dense self-supervised paradigm can be formulated as
\begin{equation}
	\mathcal{L}_{mhm}=\left\| (\mathbf{C}^{\mathbf{d}_{\tilde{\mathbf{p}}_1}}_{\mathbf{D}_{M_i}}-\mathbf{C}^{\mathbf{d}_{\tilde{\mathbf{p}}_{M_i-1}}}_{\mathbf{D}_{M_i}})_{th}\right\|_{mse}
	\label{17}
\end{equation}
Note that not all points can be correctly tracked within the patches. Therefore, it is necessary to set a threshold to filter out erroneously tracked points to avoid negatively impacting the convergence of the network.

The final self-supervise loss $\mathcal{L}_{self}$ is then a linear combination of all losses  
\begin{equation}
	\mathcal{L}_{self}=\epsilon \mathcal{L}_{srp}+\zeta \mathcal{L}_{mrp}+\eta \mathcal{L}_{mhm}
	\label{18}
\end{equation}

\section{Experiments}\label{VI}

\subsection{Datasets}\label{VI-A}
\textbf{MegaDepth} dataset \cite{8578316} provides dense depth maps and camera poses for each image, enabling the establishment of dense correspondence relationships across images. This dataset is employed to construct training frameworks for short-term downstream tasks such as Structure-from-Motion (SFM).

\textbf{HPatches} dataset \cite{8099893} includes 57 illumination-varying scenes and 59 viewpoint-varying scenes of planar images, with each scene containing five image pairs annotated with ground-truth homography matrices. Comprehensive evaluations of feature repeatability and computational efficiency are conducted on this benchmark.

\textbf{Euroc} dataset \cite{article} features six indoor and five outdoor flight sequences captured by a micro-aerial vehicle. Systematic ablation studies on VIO localization accuracy are performed using this dataset to evaluate the impact of individual modules on system performance.

\subsection{Training Details}\label{VI-B}

\textit{1) \textbf{MegaDepth training}}: FAST keypoints are detected and matched using ground-truth camera poses to ensure projected keypoints remain within the second image’s boundaries. A 64×64 pixel region centered on each feature point is cropped from both images, which serves as the network input. The loss function is formulated as \eqref{14}
\begin{equation}
\mathcal{L}_{\text{MegaDepth}}=\underbrace{\alpha \mathcal{L}_{rp }+\beta \mathcal{L}_{lpk}+\gamma \mathcal{L}_{hm}+\delta \mathcal{L}_{desc}}_{\text {Eq. }(14)}
\end{equation}

where $\alpha=1$, $\beta=0.5$, $\gamma=1$, $\delta=0.5$. The optimization process employs the ADAM optimizer \cite{Kingma2014AdamAM} with an initial learning rate of $2e^{-4}$ and batch size 2. The complete training protocol spans 50 epochs, requiring approximately 24 hours of computation on an NVIDIA GeForce RTX 4070 GPU.

\textit{2) \textbf{KITTI training}}: Temporal groups are constructed using ten consecutive frames. FAST keypoints are detected and tracked via bidirectional Lucas-Kanade (LK) optical flow, retaining only those consistently tracked across at least three consecutive frames. For each surviving feature point, a 64×64-pixel region centered at its coordinates is cropped from all frames as network input.

The composite loss function combines contributions from Equations \eqref{14} and \eqref{18}:
\begin{equation}
	\begin{split}
		\mathcal{L}_{\text{KITTI}} &= \underbrace{\alpha \mathcal{L}_{rp} + \beta \mathcal{L}_{lpk} + \gamma \mathcal{L}_{hm} + \delta \mathcal{L}_{desc}}_{\text{Eq. }(14)} \\
		&\quad + \underbrace{\epsilon \mathcal{L}_{srp} + \zeta \mathcal{L}_{mrp} + \eta \mathcal{L}_{mhm}}_{\text{Eq. }(18)}
	\end{split}
\end{equation}
with coefficients $\alpha=1$, $\beta=0.5$, $\gamma=1$, $\delta=0.5$, $\epsilon=1$, $\zeta=5$, $\eta=5$

Optimization employs the ADAM optimizer with an initial learning rate of $2e^{-4}$ and batch size 2. The complete training protocol spans 50 epochs, requiring approximately 48 hours of computation on an NVIDIA GeForce RTX 4070 GPU.

\subsection{HPatches Repeatability}\label{VI-C}

\begin{table}[htbp]
	\centering
	\caption{The MMA of various methods on original-resolution images from the HPATCHES dataset [46] at 1-5 pixel thresholds, with the top three performing results highlighted in {\color{red}\uuline{red}}, {\color{blue}\uwave{blue}}, and {\color{green}\ul{green}}, respectively.}
	\label{table:1}
	\setlength{\tabcolsep}{6pt} 
	\renewcommand{\arraystretch}{1.2} 
	\begin{tabular}{l ccc ccc}
		\toprule
		\multirow{2}{*}{\textbf{Method}} 
		& \multicolumn{3}{c}{\shortstack{\textbf{Illumination} \\ \textbf{MMA\%}}} 
		& \multicolumn{3}{c}{\shortstack{\textbf{Viewpoint} \\ \textbf{MMA\%}}} \\
		\cmidrule(lr){2-4} \cmidrule(lr){5-7}
		& \textbf{@1} & \textbf{@3} & \textbf{@5} 
		& \textbf{@1} & \textbf{@3} & \textbf{@5}\\
		\midrule
		Fast+LK         & 37.13  & 48.13  & 51.53  & 5.20  & 10.17  & 11.26\\
		Fast+LiteCNN         & \setulcolor{green}\textcolor{green}{\ul{\textbf{{53.93}}}}  & 72.13  & 81.47   & 41.44   & 57.68   & 62.79   \\
		ALIKE (N)        & 43.17  & 73.91  & 77.84 & \setulcolor{blue}\textcolor{blue}{\uwave{\textbf{50.34}}}  & \setulcolor{red}\textcolor{red}{\uuline{\textbf{76.18}}}  & \setulcolor{red}\textcolor{red}{\uuline{\textbf{80.97}}}  \\
		ZippyPoint  &   49.46 &  71.19 &  75.49 &  46.10 &  \setulcolor{green}\textcolor{green}{\ul{\textbf{69.57}}} & \setulcolor{blue}\textcolor{blue}{\uwave{\textbf{73.66}}} \\
		LiteCNN    & \setulcolor{blue}\textcolor{blue}{\uwave{{\textbf{56.98}}}} & \setulcolor{blue}\textcolor{blue}{\uwave{\textbf{{77.06}}}}  & \setulcolor{red}\textcolor{red}{\uuline{\textbf{{85.15}}}}  &  44.53  &  58.94 & 63.88 \\ 
		XFeat       &  41.25 &  \setulcolor{green}\textcolor{green}{\ul{\textbf{{76.22}}}} &  \setulcolor{blue}\textcolor{blue}{\uwave{\textbf{82.83}}} &  \setulcolor{green}\textcolor{green}{\ul{\textbf{48.62}}} &  \setulcolor{blue}\textcolor{blue}{\uwave{\textbf{70.47}}}  & 71.63  \\
		\textbf{ours}       & \setulcolor{red}\textcolor{red}{\uuline{{\textbf{58.79}}}}  & \setulcolor{red}\textcolor{red}{\uuline{\textbf{79.33}}}  & \setulcolor{green}\textcolor{green}{\ul{\textbf{{81.95}}}}  & 
		\setulcolor{red}\textcolor{red}{\uuline{\textbf{51.06}}}  & 66.24  & \setulcolor{green}\textcolor{green}{\ul{{\textbf{71.84}}}}  \\
		\bottomrule
	\end{tabular}
\end{table}

Following previous works \cite{10.5555/3454287.3455400,germain2020s2dnetlearningaccuratecorrespondences,8953622}, we evaluate the Mean Matching Accuracy (MMA) using predefined thresholds of \{1, 3, 5\} pixels. The proposed method is compared with  Fast+LK, Fast+LiteCNN, LiteCNN \cite{10556560}, ALIKE (N) \cite{9726928}, ZippyPoint \cite{10208292} and XFeat \cite{10655278}. Since the proposed approach solely focuses on the tracking process, it is embedded within the Fast feature extraction pipeline for a fair comparison. To mitigate feature point clustering, all feature detection methods employ the same non-maximum suppression (NMS) strategy. The repeatability comparison in illumination and viewpoint scenes is presented in Tab. \textcolor{red}{\ref{table:1}}. The results demonstrate that the proposed keypoint repeatability achieves a state-of-the-art performance under illumination variations. Although its performance under viewpoint changes is less competitive, this is primarily attributed to the limitations of image patches rather than the network design.

\subsection{Efficiency Comparisons with the state-of-the-arts}\label{VI-D}

The proposed method was evaluated in a PyTorch virtual environment utilizing an Intel Core i7-13700KF CPU under strictly controlled non-accelerated conditions, where both hardware acceleration (e.g., GPU/TPU) and software optimization frameworks (e.g., NCNN, ONNX) were deliberately disabled. 

\begin{figure}[h]
	\centering
	\subfigure[Single patch\label{Fig:5a}]{
		\includegraphics[width=0.45\linewidth]{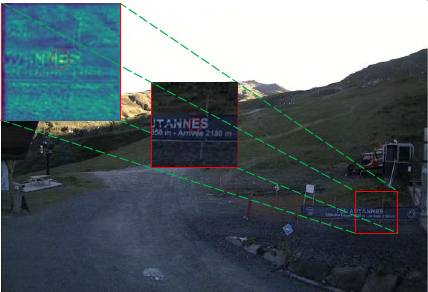}}
	\hfill
	\subfigure[Pyramid patch\label{Fig:5b}]{
		\includegraphics[width=0.45\linewidth]{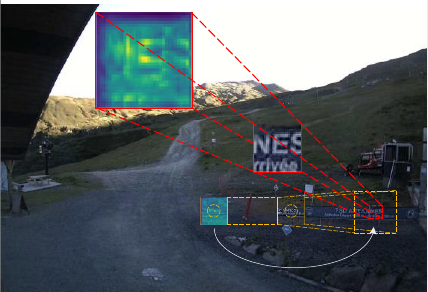}}
	\\
	\caption{Different network reasoning architectures. The left side is single-patch direct reasoning, and the right side is pyramid-patches coarse-to-fine reasoning.}
	\label{Fig:5}
\end{figure}

The proposed method is evaluated through both single-patch inference and pyramid-based inference schemes, as illustrated in Fig. \ref{Fig:5}. The patch size selection follows resolution-dependent configurations: 32×32 pixel for 480p, 64×64 pixel for 720p, 128×128 pixel for 1080p, and 256×256 pixel for 2K resolutions. The pyramid architecture employs a two-level hierarchy: Level 1 downsamples resolution-specific patches to 32×32 for coarse keypoint localization, while Level 2 processes original-resolution 32×32 pixel patches for precise keypoint positioning within identified regions.

Table II presents comprehensive complexity comparisons, including network parameters, computational complexity (GFLOPs) across resolutions, standalone inference speed, and end-to-end throughput (FPS) incorporating preprocessing (patch extraction and descriptor alignment) and postprocessing operations. Our solution achieves the most compact parameterization (9K), followed by LiteCNN (84K) and ALIKE-T (87K).  When employing single-patch inference, our method demonstrates resolution-dependent computational escalation (2857.42ms processing time) with FPS degradation to 0.12 due to postprocessing overhead, underperforming existing solutions. However, the pyramid implementation achieves sustained real-time performance (10 FPS) at high resolutions despite dual inference requirements, effectively resolving the computational bottleneck associated with high-resolution image processing. This efficiency improvement represents a 83× throughput enhancement compared to single-patch operation.

\begin{table*}[htbp]
	\centering
	\renewcommand{\arraystretch}{1.2}
	\setlength{\tabcolsep}{5pt}
	\caption{The number of network parameters, GFLOPS on images of varying resolutions, inference FPS, and end-to-end FPS, with the top three performing results highlighted in {\color{red}\uuline{red}}, {\color{blue}\uwave{blue}}, and {\color{green}\ul{green}}, respectively.}
	\begin{tabular}{l|c|ccc|ccc|ccc|ccc}
		\hline
		\textbf{Models} &  \textbf{Params/M} &\multicolumn{3}{c|}{\textbf{Low(480p)}} & \multicolumn{3}{c|}{\textbf{Medium (720p)}} & \multicolumn{3}{c|}{\textbf{High Resolution(1080p)}} & \multicolumn{3}{c}{\textbf{High Resolution(2k)}} \\
		\cline{3-14}
		&  & \textbf{GFLOPs} & \textbf{CPU/ms} & \textbf{FPS} & \textbf{GFLOPs} & \textbf{CPU/ms} & \textbf{FPS} & \textbf{GFLOPs} & \textbf{CPU/ms} & \textbf{FPS} & \textbf{GFLOPs} & \textbf{CPU/ms} & \textbf{FPS}\\
		\hline 
		ALIKE(T) & \setulcolor{green}\textcolor{green}{\ul{\textbf{{0.087}}}} & 3.40 & 151.23 & 6.24 & 10.41 & 418.23 & 2.28 & 23.09 & 971.88 & 1.01 & 40.75 & 1857.25 & 0.53 \\
		ALIKE(N) & 0.346 & 13.02 & 210.22 & 4.64 & 39.94 & 669.66 & 1.47 & 88.57 & 1502.13 & 0.65 & 156.29 & 2928.25 & 0.34 \\
		ZippyPoint & 19.85 & 3.15 & 162.58 & 1.62 & 27.96 &  610.86 & 0.83 & 62.48 & 1442.21 & 0.39 & 135.07 & 2633.72 & 0.23 \\		
		LiteCNN  & \setulcolor{blue}\textcolor{blue}{\uwave{\textbf{0.084}}} & \setulcolor{green}\textcolor{green}{\ul{\textbf{2.02}}} & 
		\setulcolor{green}\textcolor{green}{\ul{\textbf{80.62}}} & 10.36 & 
		5.92 & 
		296.41 & 3.15 &  
		13.32 &  
		\setulcolor{green}\textcolor{green}{\ul{\textbf{581.86}}} & 		\setulcolor{green}\textcolor{green}{\ul{\textbf{1.46}}} & 25.22 & 
		\setulcolor{green}\textcolor{green}{\ul{\textbf{1102.77}}} & 		\setulcolor{green}\textcolor{green}{\ul{\textbf{0.75}}} \\ 
		XFeat  &  0.658 & \setulcolor{blue}\textcolor{blue}{\uwave{\textbf{{1.33}}}} &  \setulcolor{blue}\textcolor{blue}{\uwave{\textbf{58.12}}} & 17.01 &  
		\setulcolor{green}\textcolor{green}{\ul{\textbf{3.99}}} & 
		\setulcolor{green}\textcolor{green}{\ul{\textbf{212.58}}} & 
		\setulcolor{blue}\textcolor{blue}{\uwave{\textbf{4.70}}} &  
		\setulcolor{green}\textcolor{green}{\ul{\textbf{8.96}}} & 
		\setulcolor{blue}\textcolor{blue}{\uwave{\textbf{507.88}}} & 
		\setulcolor{blue}\textcolor{blue}{\uwave{\textbf{1.97}}} & 		\setulcolor{blue}\textcolor{blue}{\uwave{\textbf{15.91}}} & 
		\setulcolor{blue}\textcolor{blue}{\uwave{\textbf{932.22}}} & 		\setulcolor{blue}\textcolor{blue}{\uwave{\textbf{1.07}}}\\
		\hline
		\multicolumn{14}{c}{\textbf{Single \& Pyramid patches}} \\
		\hline
		ours(single)  & \multirow{2}{*}{\setulcolor{red}\textcolor{red}{\uuline{\textbf{0.009}}}} & \multirow{2}{*}{\setulcolor{red}\textcolor{red}{\uuline{\textbf{0.51}}}} & 
		\multirow{2}{*}{\setulcolor{red}\textcolor{red}{\uuline{\textbf{34.46}}}} & 
		\multirow{2}{*}{\setulcolor{red}\textcolor{red}{\uuline{\textbf{19.06}}}} & 
		\setulcolor{blue}\textcolor{blue}{\uwave{\textbf{2.02}}} & 
		\setulcolor{blue}\textcolor{blue}{\uwave{\textbf{152.09}}} & 
		\setulcolor{green}\textcolor{green}{\ul{\textbf{4.83}}} &  
		\setulcolor{blue}\textcolor{blue}{\uwave{\textbf{8.14}}} & 
		586.20 & 1.28 & 		
		\setulcolor{green}\textcolor{green}{\ul{\textbf{32.39}}} & 2857.42 & 0.12 \\  
		ours(pyramid) &  &  &  &  & 
		\setulcolor{red}\textcolor{red}{\uuline{\textbf{0.51}}} & 
		\setulcolor{red}\textcolor{red}{\uuline{\textbf{35.12*2}}} & 
		\setulcolor{red}\textcolor{red}{\uuline{\textbf{10.99}}} &  
		\setulcolor{red}\textcolor{red}{\uuline{\textbf{0.51}}} & 
		\setulcolor{red}\textcolor{red}{\uuline{\textbf{35.17*2}}} & 
		\setulcolor{red}\textcolor{red}{\uuline{\textbf{10.11}}} &  
		\setulcolor{red}\textcolor{red}{\uuline{\textbf{0.51}}} & 
		\setulcolor{red}\textcolor{red}{\uuline{\textbf{36.88*2}}} & 
		\setulcolor{red}\textcolor{red}{\uuline{\textbf{9.45}}} \\
		\hline
	\end{tabular}
\end{table*}

\section{{CONCLUSION} }\label{VII}

This paper proposes a local feature point tracking method that enhances inter-frame consistency while preserving the efficiency of image patch matching, eliminating reliance on manual annotations.  The core idea of this work is to use traditional methods to assist in constraining short-term features and to use the consistency of consecutive frames to constrain long-term features. To realize this, a hybrid training mechanism is developed, combining self-supervised and unsupervised learning paradigms with a multi-loss optimization strategy. Through experiments and ablation studies on three different tasks on two different training datasets, we demonstrate the feasibility of fast and accurate image matching of high-resolution images without relying on high-level low-level hardware optimization. The results show that our method is suitable for visual localization and map-free visual relocalization, challenging downstream tasks and basic prerequisites for robotics applications, while significantly reducing the size of deep learning models, matching and localization speed.

\textbf{Acknowledgements}. The numerical computations in this study were performed on the SuperComputing System at the SuperComputing Center of Wuhan University.

\bibliographystyle{unsrt} 
\bibliography{ref}

\end{document}